\documentclass[conference]{IEEEtran}
\makeatletter
\def\ps@headings{%
\def\@oddhead{\mbox{}\scriptsize\rightmark \hfil \thepage}%
\def\@evenhead{\scriptsize\thepage \hfil \leftmark\mbox{}}%
\def\@oddfoot{}%
\def\@evenfoot{}}
\makeatother
\usepackage{cite}
\usepackage{amsmath,amssymb,amsfonts}
\usepackage{algorithmic}
\usepackage{graphicx}
\usepackage{textcomp}
\usepackage{xcolor}
\usepackage{array,multirow,graphicx}
\usepackage{pgfplots}
\usepackage{url}
\usetikzlibrary{pgfplots.groupplots}
\def\BibTeX{{\rm B\kern-.05em{\sc i\kern-.025em b}\kern-.08em
    T\kern-.1667em\lower.7ex\hbox{E}\kern-.125emX}}
\begin{document}

\title{CAN-BERT do it? Controller Area Network Intrusion Detection System based on BERT Language Model}
\author{Natasha Alkhatib, Maria Mushtaq, Hadi Ghauch, Jean-Luc Danger 
\\
\textit{Télécom Paris, IP Paris, Palaiseau, France}
\\
{\fontfamily{qcr}\selectfont
\small{\{natasha.alkhatib, maria.mushtaq, hadi.ghauch, jean-luc.danger\}@telecom-paris.fr}
}
}

\maketitle
\begin{abstract}
Due to the rising number of sophisticated customer functionalities, electronic control units (ECUs) are increasingly integrated into modern automotive systems.
However, the high connectivity between the in-vehicle and the external networks paves the way for hackers who could exploit in-vehicle network protocols' vulnerabilities.
Among these protocols, the Controller Area Network (CAN), known as the most widely used in-vehicle networking technology, lacks encryption and authentication mechanisms, making the communications delivered by distributed ECUs insecure.
Inspired by the outstanding performance of bidirectional encoder representations from transformers (BERT) for improving  many natural language processing tasks, we propose in this paper ``CAN-BERT", a deep learning based network intrusion detection system, to detect cyber attacks on CAN bus protocol.
We show that the BERT model can learn the sequence of arbitration identifiers (IDs) in the CAN bus for anomaly detection using the ``masked language model" unsupervised training objective.
The experimental results on the ``Car Hacking: Attack \& Defense Challenge 2020" dataset show that ``CAN-BERT" outperforms state-of-the-art approaches. In addition to being able to identify in-vehicle intrusions in real-time within 0.8 ms to 3 ms w.r.t CAN ID sequence length, it can also detect a wide variety of cyberattacks with an F1-score of between 0.81 and 0.99.

\end{abstract}


\section{Introduction}
To fulfill automotive features, the Controller Area Network (CAN) bus is widely used as the de-facto standard for message communication between different electronic control units (ECUs) in today's vehicles. It is sometimes referred to as a "message-based" system, since it focuses on the transmission of diagnostic, informative and controlling data through messages, also known as CAN data frames. In fact, while developing a vehicle, all conceivable CAN bus messages and their respective priority, encoded into an identifier called ``CAN ID", must be determined beforehand. 
Due to the lack of authentication, any device can connect physically or wirelessly to the CAN bus, broadcast CAN data frames, and all the participants on the CAN bus can receive it without verifying its source. Consequently, since CAN security was not a design priority, many message injection attacks have become widely implemented. These attacks can interfere with the desired function of the system, shut down some connected nodes, and make the vehicle behave abnormally, putting at risk the safety of the driver and the passengers.  
\begin{figure}
\centering
\includegraphics[scale=0.35]{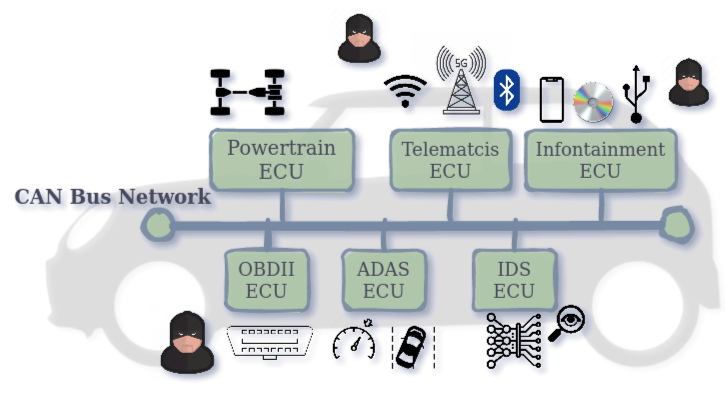}
\caption{A CAN bus network exploited by attackers through different physical and wireless interfaces.}
\label{fig:can_network}
\centering
\end{figure}
To address these security flaws, researchers have proposed intrusion detection as a supplementary layer of protection to specialized security solutions. By monitoring the communication between different ECUs within a CAN bus system, a network intrusion detection system (N-IDS) can detect deviations from the normal message exchange behavior and, thereby, identify both anticipated and novel cyberattacks.
\begin{figure*}
\centering
\includegraphics[scale=0.45]{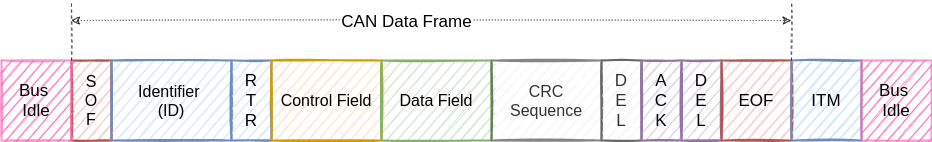}
\caption{CAN packet structure.}
\label{fig:can_frame}
\centering
\end{figure*}
The adoption of deep neural networks for in-vehicle intrusion detection have lately proliferated, with impressive results. Since a message injection attack can alter the normal order of occurence of several CAN IDs, researchers have deployed deep learning based sequential models, to model the CAN ID sequences. 
Some studies have proposed the use of autoregressive models that are trained to capture the patterns of regular CAN ID sequences by predicting the future CAN ID sequence based on the preceding one, such as recurrent neural network (RNN) models and its variants and the generative pretrained transformer (GPT). 
However, these models identify malicious network intrusions on CAN ID traffic by focusing primarily on the exchange of CAN ID messages from earlier steps rather than integrating the left and right context of a CAN ID sequence, limiting the model's capacity to grasp the whole context information representation. Additionally, these algorithms focus largely on the correlation between CAN ID messages in normal sequences, which would result in false alarms for intrusion detection whenever the correlation is breached. Hence, due to these limitations, the autoregressive models do not adequately depict the natural communication behavior between the various ECUs.

To address these challenges, we propose CAN-BERT, an intrusion detection system based on a language representation model called Bidirectional Encoder Representations from Transformers (BERT). CAN-BERT, in contrast to autoregressive models, is a self-supervised model which can successfully depict deep bidirectional representations from CAN ID sequences by conditioning on both left and right context in its various layers. By using the ``masked language model" unsupervised training objective, CAN-BERT model masks some of the CAN IDs in the input at random, with the goal of predicting the conventional ID of the masked word based on its left and right context. 

We evaluate our approach using the recently published ``Car Hacking: Attack \& Defense Challenge 2020" collected from three different cars, Chevrolet Spark, Hyundai Sonata and Kia Soul and which contains diverse types of message injection attacks. 

Our contributions are summarized below: 
\begin{itemize}
    \item  Inspired by the outstanding performance of BERT model for improving  many natural language processing tasks, we propose ``CAN-BERT", a novel BERT-based intrusion detection system architecture which can detect known and unprecedented cyberattacks in CAN ID sequences. 
    \item We evaluate the performance of our approach with the recently published ``Car Hacking: Attack \& Defense Challenge 2020" collected from three different cars, Chevrolet Spark, Hyundai Sonata and KIA Soul and which contains diverse types of message injection attacks. We also compare our model with other baseline models. 
\end{itemize}
Towards this end, our paper is organized into six sections. In Section \ref{s2}, we present an overview of the Controller Area Network (CAN) and the Bidirectional Encoder Representations from Transformers model (BERT). In Section \ref{s3}, we present an overview of our proposed framework ``CAN-BERT". Section \ref{s4} discusses the launched experiments with the corresponding dataset and the proposed metrics for IDS' evaluation. In Section \ref{s5}, we discuss the obtained results showing the proposed model accuracy and complexity. Finally, we conclude our paper with future work direction.
\section{Related Work}
Intrusion Detection systems (IDSs) have been widely used to detect intrusions on the Controller Area Network (CAN). Intrusions can be detected either by inspecting the content or the signals transmitted by CAN data frames \cite{canet}, or by examining the order by which different CAN data frames' identifiers are exchanged between the ECUs \cite{cnn}.  

\section{Preliminaries}
\label{s2}
\subsection{Controller Area Network}
The Controller Area Network (CAN), created by BOSCH in 1983, is a potent networking technology essential for the development of useful automotive features. Due to its robustness represented by its ability to allow various ECUs to be connected in almost all areas of a car, it still prevails in vehicles today. As seen in Figure, it is a bus system, meaning that all Electronic Control Units (ECUs) share the same wiring.

It is a "message-based" system in which messages, also known as CAN data frames, are transmitted between various ECUs.
As depicted in Figure. \ref{fig:can_frame}, each CAN data frame is composed of the following elements: Start of frame (SOF), identifier (ID), Remote frame transmission field (RTR), control field, data, cyclic redundancy check (CRC), delimiters (DEL) and acknowledgment fields (ACK), and end of frame fields. Each CAN bus message has a priority that is represented by the arbitration identifier field (ID) that can either be composed of 11 or 29 bits, depending on the car manufacturer and which will be mainly used in our work for detecting in-vehicle intrusions.

To avoid contention between multiple ECUs willing to transmit CAN messages in the medium, CAN employs a priority-based mechanism which allows the ECU with the highest priority/lowest value identifier to transmit before others. The procedure is termed ``arbitration" because the message with the highest priority wins out over competing messages with a lower priority at the time of transmission. 

\begin{figure}
\centering
\includegraphics[scale=0.42]{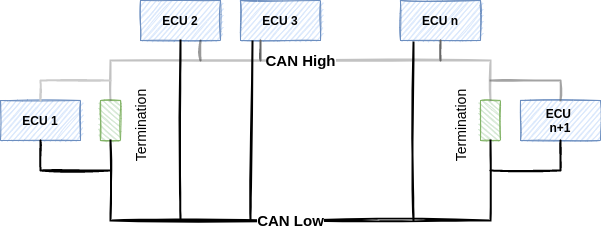}
\caption{ECU devices connected through CAN bus, source \cite{AutoEth_book}.}
\label{fig:can_topology}
\centering
\end{figure}
\subsubsection{Security Weaknesses}

CAN does not prohibit several ECUs from sharing the same IDs. Moreover, CAN messages are broadcast and do not contain any sender's address. Consequently, any device linked to the CAN bus can use any pre-defined ID, communicate its message without authentication or encryption, and all associated ECUs can receive it. The receiver defines whether or not a message identification causes the receiving ECU to retain and process the given data. Consequently, an attacker is able to broadcast spoofed CAN messages, eavesdrop on all the CAN traffic and collect detailed information about it, resulting in Fuzzing and Malfunction attacks. 

Additionally, as previously mentionned, the CAN bus leverages the arbitration method which discerns between "dominant" (0) and "recessive" (1) bits in the message identifiers. Therefore, if several ECUs begin transmitting simultaneously, the ECU whose message begins with a greater number of dominant "0" bits will take over the CAN bus. As soon as a unit detects that the message on the bus is no longer the message it is transmitting, it halts its transmission, waits for the real transmission to conclude, then waits for the interframe gap to expire and retransmits its message. This phenomenon carries the risk that a message with a lower priority will never be delivered if the network is very congested and can be exploited by attackers to launch denial of service (DoS)/ flooding attacks. 
\subsection{BERT}
\begin{figure*}[h]
\centering
\includegraphics[scale=0.35]{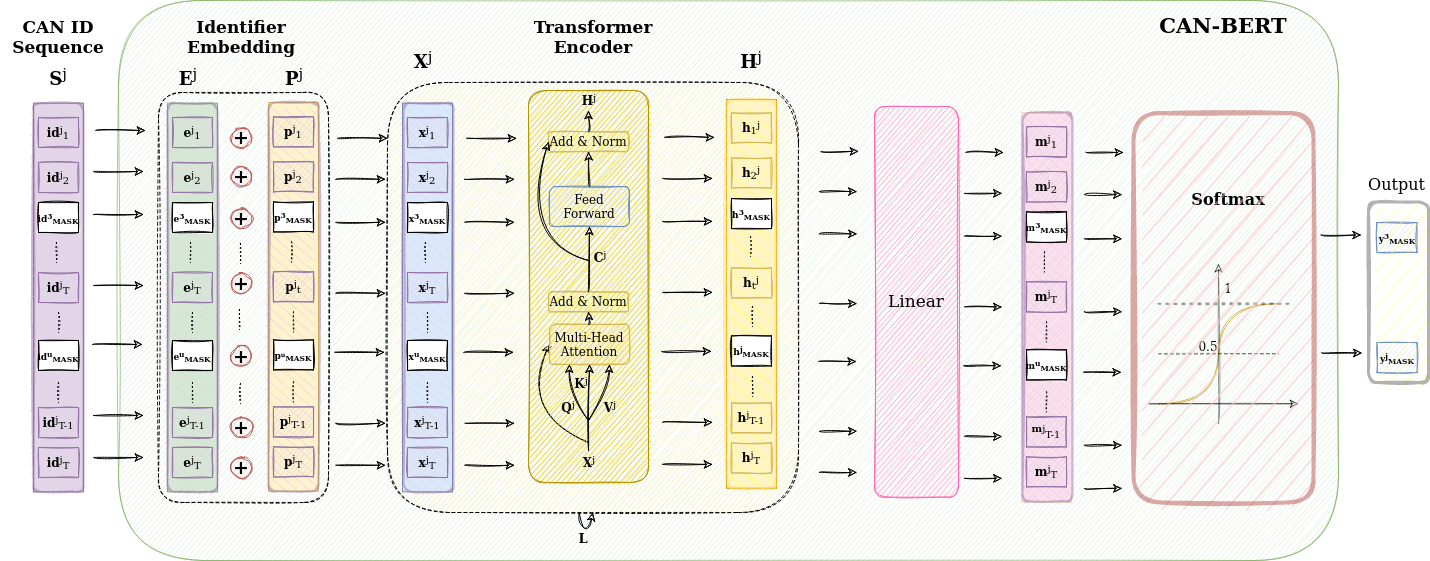}
\caption{CAN-BERT model architecture}
\label{fig:can-bert}
\centering
\end{figure*}

Bidirectional Encoder Representations from Transformers (BERT), proposed by Devlin et al. \cite{BERT}, is a state-of-the-art language representation model which is designed to pretrain bidirectional representations from unlabeled text by jointly conditioning on both left and right context in all layers. Regarding its architecture, it is a multi-layer bidirectional Transformer encoder based on the original implementation proposed by Vaswani et al. \cite{Transformer}. 

Inspired by its outstanding performance in modeling sequential data, BERT is recently employed for sequence anomaly detection \cite{Language_anomaly_bert} \cite{LogBERT} \cite{LAnoBERT} \cite{NeuralLog} \cite{IOT_BERT}. To the best of our knowledge, none of the previous works have evaluated the performance of the BERT model for in-vehicle intrusion detection on CAN protocol.

In order to detect anomalies in sequences, it's crucial to incorporate context from both left and right directions of the sequence. Sequential anomalies may be misdetected by traditional unidirectional models, such as OpenAI GPT and RNNs, where every token can only attend to context to its left. To solve this significant constraint, some researchers have proposed a shallow concatenation of both left-to-right and right-to-left architecture of the autoregressive models , such as Bi-RNN and Bi-GPT \cite{Bi-GPT}. However, these approaches aren't as powerful as BERT which adopts a "masked language model" (MLM) training objective, in which input sequence tokens are randomly masked and the goal is to predict the original vocabulary id of the masked word based on its context. In contrast to denoising auto-encoders, BERT predicts the masked words instead of reconstructing the whole sequence \cite{AE}. 

\section{Proposed framework: CAN-BERT}
\label{s3}
\label{proposed_canbert}
We propose "CAN-BERT", a pattern-based anomaly detection algorithmn, which leverages a BERT-based architecture to detect message injection intrusions in the CAN bus. As seen in Figure. \ref{fig:can-bert}, our model is composed of a multi-layer bidirectional Transformer encoder and is trained using the ''masked language model" self-supervised task to model normal CAN ID sequences. The following subsections elaborately describe the suggested framework.

\subsection{Model description}
Note that $\mathbf{S}=[ \mathbf{id}_{1},..., \mathbf{id}_{t},...,\mathbf{id}_{T}]$ is an observed sequence of $T$ CAN identifiers, arranged in their order of transmission in the CAN bus network, where an identifier  $\mathbf{id}_{t} \in \mathbb{ID}$ is an M-dimensional vector which denotes the CAN ID transmitted at time $t$ by an ECU, $\mathbb{ID}$ indicates the set of CAN IDs extracted from CAN messages, and M is the size of the $\mathbb{ID}$ set. 

Since anomaly detection is an unsupervised learning-based technique in which only normal data are used for training, a collection of $N$ CAN ID sequences, represented as $\mathbb{D}_{training}=\{ \mathbf{S}_{1},..., \mathbf{S}_{j},...,\mathbf{S}_{N}\}$, is solely used as a training dataset.  

\noindent \textbf{Identifier Embeddings} To feed the appropriate input to the BERT model, each identifier $\mathbf{id}^{j}_{t}$ with size (M,1) in a CAN sequence $\mathbf{S}^{j}$ is firstly projected into a $d$-dimensional space using a single linear layer, i.e.: 
\begin{equation}
    \mathbf{e^{j}_{t}}=\mathbf{W}^{e}\mathbf{id^{j}_{t}}+\mathbf{b}^{e}, \forall i \in \{1..T\},  \forall j \in \{1..N\}
\end{equation}
where $\mathbf{e^{j}_{t}}$ represents the identifier embedding with size ($d$,1), $\mathbf{W}^{e} \in \mathbb{R}^{d \times M}$ is the input embedding weight matrix, and $\mathbf{b}^{e} \in \mathbb{R}^{d}$ denotes the bias. Both $\mathbf{W}^{e}$ and $\mathbf{b}^{e}$ are trainable parameters.

Subsequently, the identifier's position is encoded into a $d$-dimensional positional embedding $\mathbf{p^{j}_{t}}$ using a sinusoidal function. To this end, the CAN ID, fed into the CAN-BERT model, is a summation of both the positional encoding and the input embedding : 
\begin{equation}
\label{input}
\mathbf{x}^{j}_{t}=\mathbf{e^{j}_{t}}+\mathbf{p^{j}_{t}}
\end{equation}
where $\mathbf{x}^{j}_{t}$ is the total embedding $j$-th identifier in the $t$-th CAN ID sequence $\mathbf{id}^{j}_{t}$, thereby the convergence of the input sequence $\mathbf{S}^{j}$ into $\mathbf{X}^{j}=[\mathbf{x}^{j}_{1},..\mathbf{x}^{j}_{t}..,\mathbf{x}^{j}_{T}]^{T}$ with $\mathbf{X}^{j}$ a matrix with size $T \times d$.

\noindent \textbf{Transformer Encoder} The encoded input $\mathbf{X}^{j}$ is then delivered into a stack of $L$ transformer encoder layers, each of which has two sub-layers: a multi-head self-attention mechanism and a position-wise feed-forward network \cite{Transformer}. A residual connection is employed around each of these two sub-layers, followed by layer normalization \cite{Normalization}, as follows: 
\begin{equation}
\label{sublayers}
\begin{split}
        \mathbf{E}^{(j,l)}=g(\mathbf{X}^{(j,l)}) + f(\mathbf{X}^{(j,l)}+g(\mathbf{X}^{(j,l)})) \\
        \mathbf{H}^{(j,l)}=z(\mathbf{E}^{(j,l)}) + f(\mathbf{E}^{(j,l)}+z(\mathbf{E}^{(j,l)})) \\
        \mathbf{X}^{(j,l+1)}=\mathbf{H}^{(j,l)},  \forall l < L
\end{split}
\end{equation}
where $\mathbf{E}^{(j,l)}$ represents the output of the first sublayer for the $l$-th transformer encoder layer with size $T \times d$, $\mathbf{H}^{(j,l)}$ represents the output of the second sublayer for the $l$-th transformer encoder layer with size $T \times d$, $g$ is the multi-head attention function, $z$ is the position wise feed forward function, and $f$ is the layer normalization function.

\textbf{Attention} We use the scaled dot-product attention proposed by \cite{Transformer}, requiring query $\mathbf{Q}^{(j,l)}$, key $\mathbf{K}^{(j,l)}$, and value $\mathbf{V}^{(j,l)}$ representations, and which are projections of the embedded sequence $\mathbf{X}^{(j,l)} \in \mathbb{R}^{T \times d}$. In fact, we leverage the dot-product similarity to compare the query representation of a given CAN identifier to all other keys. Hence, if the query and key are comparable have a high attention weight, the matching value is deemed to be relevant. The output is therefore computed as a weighted sum
of the values $\mathbf{V}$:
\begin{equation}
\begin{split}
     Attn(\mathbf{Q}^{(j,l)},\mathbf{K}^{(j,l)},\mathbf{V}^{(j,l)})=\sigma(\frac{\mathbf{Q}^{(j,l)}\mathbf{K}^{(j,l)T}}{\sqrt{d}})\mathbf{V^{(j,l)}} \\
     Attn(\mathbf{Q}^{(j,l)},\mathbf{K}^{(j,l)},\mathbf{V}^{(j,l)})=\mathbf{A}\mathbf{V}^{(j,l)}
\end{split}
\end{equation}
where $\sigma$ is the softmax function, $\mathbf{A} \in \mathbb{R}^{T \times T}$ denotes the attention weight matrix containing attention weights, and $d$ is the dimension of the $\mathbf{Q}^{(j,l)}$, $\mathbf{K}^{(j,l)}$,$\mathbf{V}^{(j,l)}$ vectors.

 As described by \cite{Transformer}, the multiple heads of attention allows the model to concurrently capture diverse aspect of data at distinct CAN IDs. Hence, we adopt a multi-head attention (MHA) mechanism in which the $d$-dimensional CAN identifers are projected into subspaces calculated by different attention heads $n \in \{1,..,H\}$:
\begin{equation}
\begin{split}
    \mathbf{Q}^{(j,n,l)}= \mathbf{X}^{(j,l)}\mathbf{W}^{(Q,n)}, \mathbf{Q}^{(j,n,l)} \in \mathbb{R}^{T \times F} \\
    \mathbf{K}^{(j,n,l)}= \mathbf{X}^{(j,l)}\mathbf{W}^{(K,n)}, \mathbf{K}^{(j,n,l)} \in \mathbb{R}^{T \times F} \\
    \mathbf{V}^{(j,n,l)}= \mathbf{X}^{(j,l)}\mathbf{W}^{(V,n)}, \mathbf{V}^{(j,n,l)} \in \mathbb{R}^{T \times F}
\end{split}
\end{equation}
where $\mathbf{Q}^{(j,n,l)}$, $\mathbf{K}^{(j,n,l)}$ and $\mathbf{V}^{(j,n,l)}$ are the query, key and value vectors, respectively of the $j-th$ CAN ID sequence for the $l$-th transformer encoder layer and which are calculated using the $n$-th attention head. The $\mathbf{W}^{(Q,n)}$, $\mathbf{W}^{(K,n)}$ and $\mathbf{W}^{(V,n)}$ are their corresponding trainable weight matrices $\in \mathbb{R}^{d \times F}$, and $F$ is set to $D/H$. The results are then concatenated and projected back into representation space using the weight matrix $\mathbf{W}^{o} \in \mathbb{R}^{HF \times D}$ as follows:
\begin{equation}
\label{attention_head}
    \mathbf{head}^{(j,l)}_{n}=Attn(\mathbf{Q}^{(j,n,l)},\mathbf{K}^{(j,n,l)},\mathbf{V}^{(j,n,l)})
\end{equation}
\begin{equation}
\label{concat_head}
    \overline{\mathbf{X}}^{(j,l)}=[\mathbf{head}^{(j,l)}_{1},..\mathbf{head}^{(j,l)}_{n},..,\mathbf{head}^{(j,l)}_{H}]\textbf{W}^{O}
\end{equation}
where $\overline{\mathbf{X}}^{(j,l)}  \in \mathbb{R}^{(T,d)}$.

\textbf{Position-wise feed-forward} A position-wise feed-forward network with a ReLU activation is thereby applied to each representation in each of the layers of our encoder, in addition to attention sub-layers, using the following equation: 
\begin{equation}
\label{ffn}
    z(\mathbf{E}^{(j,l)})=[\mathbf{W_{1}}\mathbf{E}^{(j,l)} ]_{+} \circ \mathbf{W_{2}}
\end{equation}
where  $\mathbf{E}^{(j,l)}$ is previously defined in (\ref{sublayers}), $\textbf{W}_{1}$ and $\textbf{W}_{2}$ are trainable projection matrices, where $\circ$ is the hadarmard product, and $[ ]_{+}$ is the element-wise maximum. 

After passing through different transformer layers, the $L$-th contextual embedding vectors of the CAN IDs, denoted as $\mathbf{h}^{(j,L)}_{t}$ with size $(d,1) \in \mathbf{H}^{(j,L)}=[\mathbf{h}^{(j,L)}_{1},..,\mathbf{h}^{(j,L)}_{T}]^{T}$, are fed into a single linear layer which projects them back to the $M$-dimensional layer, as follows: 
\begin{equation}
    \mathbf{m^{j}_{t}}=\mathbf{W}^{m}\mathbf{h}^{(j,L)}_{t}+\mathbf{b}^{m}, \forall i \in \{1..T\},  \forall j \in \{1..N\}
\end{equation}
where $\mathbf{m^{j}_{t}}$ represents the projected output with size ($M$,1), $\mathbf{W}^{m} \in \mathbb{R}^{M \times d}$ is the input embedding weight matrix, and $\mathbf{b}^{e} \in \mathbb{R}^{M}$ denotes the bias. Both $\mathbf{W}^{m}$ and $\mathbf{b}^{m}$ are trainable parameters.
\subsection{Training and Inference}
We use the masked language model training method to train the CAN-BERT model on capturing the patterns of normal CAN ID sequences. Hence, CAN sequences with random mask as inputs, where we randomly replace a ratio of CAN IDs in a sequence with a specific MASK token, are fed into CAN-BERT. The purpose of training is to reliably anticipate the CAN IDs that have been randomly masked.

To achieve that, we feed the contextual embedding vector of the $u$-th MASK in the j-th CAN ID sequence $\textbf{m}^{j}_{MASK_{u}}$ to a softmax function, which will return a probability distribution for the whole set of CAN IDs $\mathbb{ID}$: 
\begin{equation}
\label{output}
    \hat{\mathbf{y}}^{j}_{[MASK_{u}]}=\sigma(\mathbf{W}^{c} \mathbf{m}^{j}_{[MASK_{u}]}+\mathbf{b}^{c})
\end{equation}
where $\hat{\mathbf{y}}^{j}_{[MASK_{u}]}$ is an $m$-dimensional vector, $\sigma$ is the softmax function, $\mathbf{m}^{j}_{[MASK_{u}]}$ and $\mathbf{b}^{c}$ are trainable parameters. 

CAN-BERT is trained to minimize the cross entropy loss over a batch of $I$ sequences ( with $ I \leq N$), for masked CAN ID prediction, which is defined as: 
\begin{equation}
\label{objective_function}
    \mathcal{L}_{MASK}=-\frac{1}{IR}\sum^{N}_{j=1}\sum^{R}_{u=1}{\mathbf{y}}^{j}_{[MASK_{u}]}log\hat{\mathbf{y}}^{j}_{[MASK_{u}]}
\end{equation}
where the ground-truth $u$-th CAN ID in the $j$-th sequence is denoted as ${\mathbf{y}}^{j}_{[MASK_{u}]}$, $R$ is the total number of masked tokens in the $j$-th sequence, and $N$ is the number of training samples. 

By modeling the normal exchange of messages through CAN bus using CAN-BERT, our model is expected, after training, to predict a candidate set with the normal CAN IDs having the highest likelihood for each masked token. Hence, for a randomly masked testing sequence, we calculate the probability distribution represented in (\ref{output}), for each masked CAN ID. Therefore, if the actual CAN ID is among the anticipated candidates, the corresponding CAN ID sequence is considered as normal. Otherwise, it is deemed abnormal. 
\section{Experimental Settings}
 \label{s4}
\subsection{Dataset}
To assess the proposed CAN-BERT, we leverage the ``In-Vehicle Network Intrusion Detection Challenge" dataset \cite{Dataset} (presented in Table \ref{table:ivnid_dataset}), which was used at the ``In-vehicle Network Intrusion Detection track’ of ‘Information Security R\&D Data Challenge 2019. It includes normal and abnormal in-vehicle network traffic data of HYUNDAI Sonata, KIA Soul, and CHEVROLET Spark vehicles collected during their stationary state. We have mainly used its preliminary dataset, which includes three types of attacks (Flooding, Fuzzy, and Malfunction). The dataset is labeled and each sample is represented by the following features: ``Timestamp" representing the logging time, ``CAN ID" representing the CAN Identifier, ``DLC" indicating the Data length code, and the Payload indicating the ``CAN data" field.
\subsubsection{Attacks}
The dataset contains the following attacks: 
\begin{itemize}
    \item \textbf{Flooding Attack} The flooding attack was carried out by injecting many messages with the CAN ID set to 0x000 into the CAN network. Consequently, an ECU that transmits CAN data frames with such CAN ID dominates the CAN bus, which could restrict the communications among the ECU nodes and impair normal in-vehicle functions.
    \item \textbf{Fuzzy Attack} To implement fuzzy attacks, the attacker injected every 0.0003 seconds random CAN packets, for both the ID field and the Data field. This process lead to abnormal automotive functionalities behavior such as short beeping sound repeatedly occurring, the heater turning on, etc.
    \item \textbf{Malfunction Attack} The malfunction attack was carried out by injecting messages with a specified CAN ID from among the extractable CAN IDs of a particular vehicle in order to disable relevant automotive functions, such as IDs 0×316, 0×153 and 0×18E for the HYUNDAI YF Sonata, KIA Soul, and CHEVROLET Spark vehicles, respectively.
\end{itemize}
\begin{figure*}[!h]
\centering
\begin{tikzpicture}
  \begin{groupplot}[group style={group size=4 by 1},
  	height=0.2\textheight,width=0.25\textwidth]
    
\nextgroupplot[ymin=70,ymax=100,ylabel={F1-Score}, xtick={1,...,4},ytick pos=left,title={$m= 0.15$},xticklabels={1,2,4,8},xlabel=$h$]
\addplot[
    color=red,
    mark=x,
    ]
    coordinates { 
  (1,90.12)
  (2,90.74)
  (3,90.91)
  (4,91.55)
    };
\addplot[
    color=blue,
    mark=x,
    ]
    coordinates { 
  (1,78.41)
  (2,79.18)
  (3,79.44)
  (4,80.27)  
    };
\addplot[
    color=magenta,
    mark=x,
    ]
    coordinates { 
   (1,74.45)
  (2,71.89)
  (3,73.13)
  (4,72.50)
    };
\nextgroupplot[ymin=70,ymax=100,title={$m= 0.3$},xtick={1,...,4},xticklabels={1,2,4,8},xlabel=$h$]
\addplot[
    color=red,
    mark=x,
    ]
    coordinates { 
(1,95.56)
(2,95.62)
(3,96.25)
(4,96.50)
    };
\addplot[
    color=blue,
    mark=x,
    ]
    coordinates { 
(1,86.48)
(2,86.46)
(3,86.36)
(4,86.71)
    };
\addplot[
    color=magenta,
    mark=x,
    ]
    coordinates { 
(1,80.22) 
(2,78.95)
(3,77.95)
(4,77.42)

    };
\nextgroupplot[ymin=70,ymax=100,title={$m= 0.45$},xtick={1,...,4},xticklabels={1,2,4,8},xlabel=$h$]
\addplot[
    color=red,
    mark=x,
    ]
    coordinates { 
  (1,97.06)
  (2,97.46)
  (3,97.61)
  (4,97.09)

    };
\addplot[
    color=blue,
    mark=x,
    ]
    coordinates { 
  (1,86.67)
  (2,87.59)
  (3,87.77)
  (4,87.39)

    };
\addplot[
    color=magenta,
    mark=x,
    ]
    coordinates { 
  (1,84.87)
  (2,83.16)
  (3,83.11)
  (4,82.22)
    };

\nextgroupplot[ymin=70,ymax=100,title={$m= 0.6$},xtick={1,...,4}, xlabel=$h$, xticklabels={1,2,4,8},legend style={at={($(-0.5,0)-((3cm,1.4cm)$)},legend
        columns=3,fill=none,draw=black,anchor=center,align=center}]
\addplot[
    color=red,
    mark=x,
    ]
    coordinates { 
(1,95.60)
(2,96.86)
(3,96.87)
(4,96.91)
    };
\addplot[
    color=blue,
    mark=x,
    ]
    coordinates { 
(1,85.12)
(2,85.21)
(3,85.58)
(4,85.97)
    };
\addplot[
    color=magenta,
    mark=x,
    ]
    coordinates { 
(1,78.79)
(2,79.77)
(3,77.60)
(4,78.29)
    };

\addlegendentry{Flooding};  
\addlegendentry{Fuzzy}; 
\addlegendentry{Malfunction}; 

\end{groupplot}
\end{tikzpicture}

\caption{{Hyperparameter tuning the mask ratio $m$ and the number of attention heads $h$} on the ``CHEVROLET Spark" dataset for $T$=32. We have obtained similar behavior pattern for the results w.r.t other sequence length and car types. }
\label{hyperparameters}
\end{figure*}
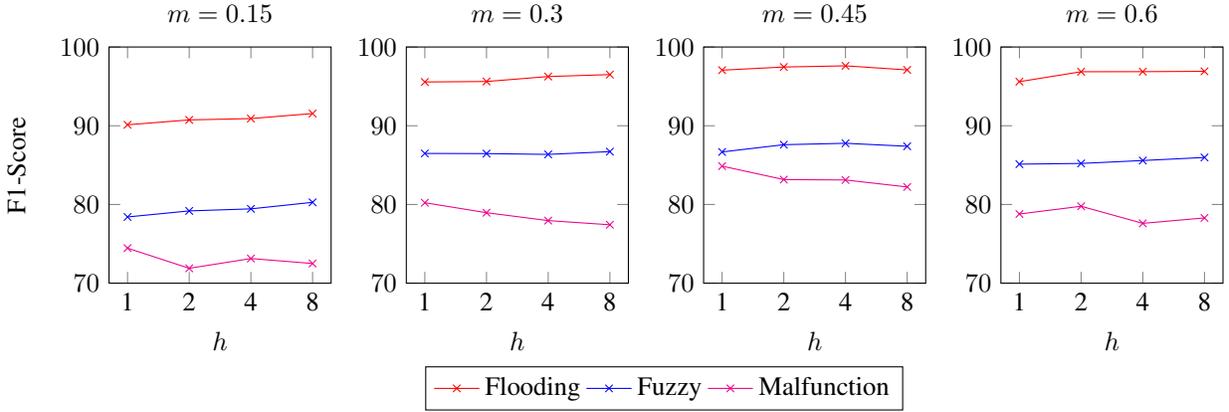
\begin{table}[h]
\begin{center}
\caption{In-Vehicle Network Intrusion Detection Dataset \label{table:ivnid_dataset}}
\centering
\begin{tabular}{|l | l | l | l| l|} 
\hline
\multirow{2}{*}{\textbf{Vehicle}} & \textbf{Dataset} & \textbf{\# Normal} & \textbf{\# Abnormal} & \textbf{Size} \\
& & \textbf{ packets} & \textbf{packets} & \textbf{(MB)} \\
\hline
 & Attack Free & 136,933 & N/A & 6.2\\
                                 \cline{2-5}
            CHEVROLET                     & Flooding & 70,001 & 14,999 & 4.2\\
                                 \cline{2-5}
              Spark                   & Fuzzy & 37,957 & 3,043 & 2.0\\
                                 \cline{2-5}
                                 & Malfunction & 47,005 & 3,995 & 2.5 \\
\hline
 & Attack Free & 117,172 & N/A & 5.8 \\
                                 \cline{2-5}
               HYUNDAI                  & Flooding & 78,907 & 17,093 & 4.9\\
                                 \cline{2-5}
                   Sonata              & Fuzzy & 78,905 & 9,095 & 4.5\\
                                 \cline{2-5}
                                 & Malfunction & 78,798 & 8,202 & 4.5\\
\hline
 & Attack Free & 192,515 & N/A & 9.3 \\
                                 \cline{2-5}
                     KIA            & Flooding & 103,928 & 16,072 & 6.2\\
                                 \cline{2-5}
                      Soul           & Fuzzy & 122,387 & 21,613 & 7.4\\
                                 \cline{2-5}
                                 & Malfunction & 108,230 & 4,770 & 5.8\\
\hline
\end{tabular}
\end{center}
\end{table}

As mentioned in Section \ref{proposed_canbert}, we aim to detect if a sequence of ordered CAN ID contains injected messages. Hence, in order to represent CAN ID sequences, we use the \textbf{Feature-based Sliding Window (FSW)} to group CAN IDs which belong to the dataset into subsequences with fixed window size $\mathbf{T}$, where $\mathbf{T}$ $\in \{16,32,64,128,256\}$ and the slide size is 1. Moreover, each CAN ID sequence $\mathbf{S}=[ \mathbf{id}_{1},..., \mathbf{id}_{t},...,\mathbf{id}_{T}]$ has its corresponding labels represented by $\mathbf{Y}=[ \mathbf{y}_{1},..., \mathbf{y}_{t},...,\mathbf{y}_{T}]$ wherein each identifier $\mathbf{id}_{t} \in \mathbf{S} $ is labeled as $\mathbf{y}_{t}=1$ if $\mathbf{id}_{t} $ is an injected identifier in $\mathbf{S}$ or as $\mathbf{y}_{t}=0$ otherwise. However, to identify the state of each sequence, we have used the following criteria: 
\begin{equation*}
 z = \begin{cases} \mbox{1 (abnormal)} & \textup{if }  \exists \mathbf{y}_{t}=1 , \forall t \in \{1, .., T\} \nonumber \\ 
                      \mbox{0 (normal)} & \textup{otherwise}
                \end{cases} 
\end{equation*}
where $z$ is the CAN ID sequence's label. 

\subsection{Benchmark Models}
The benchmark models for evaluating the performance of different CAN ID sequence anomaly detection algorithms with CAN-BERT on the chosen dataset, are detailed in this section.
\begin{itemize}
    \item \textbf{Principal Component Analysis (PCA):} PCA \cite{PCA} is a feature selection model which can be used to reduce data features from $m$ dimensions to $n$. Inverting the PCA transform does not retrieve the data lost during the application of the transform. The essence of PCA-based anomaly identification is that an anomalous sample should have more loss or reconstruction error than a normal sample. In other words, the loss sustained when an anomalous sample is processed by a PCA algorithm and projected back to its dimension using PCA also should be greater than when the same procedure is performed on a normal sample.
    \item \textbf{Isolation Forest (iForest):} Isolation forest (IF), proposed by Liu at al. \cite{IsolationForest}, detects anomalies using isolation rather than modelling the normal points. In fact, this technique presents a novel approach for isolating anomalies using binary trees, providing a new prospect for a speedier anomaly detector that directly targets abnormalities rather than profiling all regular instances.
    \item \textbf{Autoencoder (AE):}
     The autoencoder, introduced by Rumelhart et al. \cite{AE1}, is a deep learning based algorithm which seeks to learn a low-dimensional feature representation space suitable for reconstructing the provided data instances. During the encoding process, the encoder maps the original data onto low-dimensional feature space, while the decoder tries to retrieve the original data from the projected low-dimensional space. Reconstruction loss functions are used to learn the parameters of the encoder and decoder networks. Its reconstruction error value must be minimized during the training of normal instances and therefore used during testing as an anomaly score. In other words, compared to the typical data reconstruction error, anomalies that differ from the majority of the data have a large data reconstruction error. In our experiments, we have tested Long short-term based memory (LSTM) and Bidirectional LSTM (BiLSTM) models with different network hyperparameters: BiLSTM-AE-4 (with 4 layers), LSTM-AE-4 (with 4 layers), and LSTM-AE-8 (with 8 layers).
\end{itemize}
\subsection{Evaluation metrics}
\label{evaluation_metrics}
For measuring the performance of different anomaly-based IDS, we use the F1-score metric, a weighted average result of both metrics precision and recall and which is specifically used when the dataset is imbalanced. The model has a large predictive power if the F1-score is near 1.0.

Precision is the ratio of correctly classified predicted abnormal observations of all the observations in the predicted class.
\begin{equation}
    Precision = \frac{TP}{TP + FP}
\end{equation}
Recall is the ratio of correctly predicted abnormal observations of all observations in the actual class.
\begin{equation}
    Recall = \frac{TP}{TP + FN}
\end{equation}
Hence, the F1-score is calculated using the following equation:
\begin{equation}
    F1-score= 2 \cdot \frac{Precision \cdot Recall}{Precision + Recall}
\end{equation}

Where: TP= True Positive; FP=False Positive; TN= True Negative; FN=False Negative.

\begin{figure*}[!h]
\centering
\begin{tikzpicture}
  \begin{groupplot}[group style={group size=3 by 3},
  					height=0.2\textheight,width=0.3\textwidth]
    
\nextgroupplot[xtick={1,...,5},title=CHEVROLET Spark,ylabel={Flooding}, ytick pos=left,xticklabels={16,32,64,128,256}]
\addplot[
    color=red,
    mark=x,
    ]
    coordinates { 
  (1,85.59) 
  (2,97.09)
  (3,99.69) 
  (4,99.75)
  (5,99.69)

    };

\addplot[
    color=magenta,
    mark=x,
    ]
    coordinates { 
  (1,20.80)
  (2,20.15)
  (3,21.31)
  (4,21.88)
  (5,20.15) 

    };
\addplot[
    color=green,
    mark=x,
    ]
    coordinates { 
  (1,11.82)
  (2,11.73)
  (3,11.72)
  (4,11.74)
  (5,11.82) 
    };
\addplot[
    color=blue,
    mark=x,
    ]
    coordinates { 
(1,74.47)
(2,64.48)
(3,62.43)
(4,61.89)
(5,62.73)
    };
\addplot[
    color=orange,
    mark=x,
    ]
    coordinates { 
(1,69.10)
(2,67.17)
(3,63.82)
(4,62.16)
(5,88.97)
    };
\addplot[
    color=cyan,
    mark=x,
    ]
    coordinates { 
(1,96.39)
(2,89.76)
(3,90.44)
(4,84.30)
(5,62.24)
    };
\nextgroupplot[xtick={1,...,5},title=HYUNDAI Sonata,xticklabels={16,32,64,128,256}]
\addplot[
    color=red,
    mark=x,
    ]
    coordinates {
  (1,91.20)
  (2,99.59)
  (3,99.69)
  (4,99.82)
  (5,99.83)
    };
\addplot[
    color=magenta,
    mark=x,
    ]
    coordinates {
  (1,24.11)
  (2,24.04)
  (3,24.13)
  (4,23.18)
  (5,24.11)
    };
\addplot[
    color=green,
    mark=x,
    ]
    coordinates { 
  (1,11.23)
  (2,11.14)
  (3,11.03)
  (4,11.10)
  (5,11.23)
    };
\addplot[
    color=blue,
    mark=x,
    ]
    coordinates { 
(1,66.86)
(2,67.72)
(3,60.01)
(4,57.25)
(5,55.17)
    };
\addplot[
    color=orange,
    mark=x,
    ]
    coordinates { 
(1,71.15)
(2,65.38)
(3,61.05)
(4,55.94)
(5,0)
    };
\addplot[
    color=cyan,
    mark=x,
    ]
    coordinates { 
(1,90.01)
(2,68.93)
(3,85.15)
(4,57.14)
(5,57.75)
    };
\nextgroupplot[xtick={1,...,5},title=KIA Soul,xticklabels={16,32,64,128,256}]
\addplot[
    color=red,
    mark=x,
    ]
    coordinates {
  (1,81.81)
  (2,97.89)
  (3,99.77)
  (4,99.73)
  (5,99.65)
    };
\addplot[
    color=magenta,
    mark=x,
    ]
    coordinates {
  (1,18.13)
  (2,18.18)
  (3,18.73)
  (4,18.85)
  (5,18)
    };
\addplot[
    color=green,
    mark=x,
    ]
    coordinates { 
  (1,13.10)
  (2,13.13)
  (3,13.19)
  (4,13.25)
  (5,13)
    };
\addplot[
    color=blue,
    mark=x,
    ]
    coordinates { 
(1,48.61)
(2,47.19)
(3,47.28)
(4,40.55)
(5,1.079)
    };
\addplot[
    color=orange,
    mark=x,
    ]
    coordinates { 
(1,51.33)
(2,47.37)
(3,47.27)
(4,33.30)
(5,0)
    };
\addplot[
    color=cyan,
    mark=x,
    ]
    coordinates { 
(1,68.95)
(2,92.75)
(3,47.26)
(4,47.99)
(5,47.83)
    };
\nextgroupplot[xtick={1,...,5},ylabel={Fuzzing}, ytick pos=left,xticklabels={16,32,64,128,256}]
\addplot[
    color=red,
    mark=x,
    ]
    coordinates {
  (1,74.62)
  (2,87.39)
  (3,96.58) 
  (4,99.39)
  (5,99.35)
    };
\addplot[
    color=magenta,
    mark=x,
    ]
    coordinates {
  (1,29.421)
  (2,29.475)
  (3,29.94)
  (4,29.20)
  (5,29.42) 
    };
\addplot[
    color=green,
    mark=x,
    ]
    coordinates {
  (1,26.59)
  (2,26.62)
  (3,26.67)
  (4,26.63)
  (5,26.59) 
    };
\addplot[
color=blue,
mark=x,
]
coordinates {
(1,50.42)
(2,56.14)
(3,54.92)
(4,48.23)
(5,79.42)
};
\addplot[
color=orange,
mark=x,
]
coordinates {
(1,52.33)
(2,56.48)
(3,49.93)
(4,71.40)
(5,88.97)
};
\addplot[
color=cyan,
mark=x,
]
coordinates {
(1,77.04)
(2,76.46)
(3,72.93)
(4,76.14)
(5,71.15)
};
\nextgroupplot[xtick={1,...,5},xticklabels={16,32,64,128,256}]
\addplot[
    color=red,
    mark=x,
    ]
    coordinates {
  (1,70.28)
  (2,93.65) 
  (3,98.87)
  (4,99.58)
  (5,99.21)
    };
    
\addplot[
    color=magenta,
    mark=x,
    ]
    coordinates {
 (1,37.271)
 (2,37.292)
 (3,37.29)
 (4,37.01)
 (5,37.55)
    };
\addplot[
    color=green,
    mark=x,
    ]
    coordinates {
  (1,36.179)
  (2,36.246)
  (3,36.25)
  (4,36.18)
  (5,36.10)
    };
\addplot[
color=blue,
mark=x,
]
coordinates {
(1,54.70)
(2,55.50)
(3,50.29)
(4,47.90)
(5,50.23)
};
\addplot[
    color=orange,
    mark=x,
    ]
    coordinates { 
(1,60.50)
(2,55.55)
(3,50.97)
(4,51.90)
(5,61.16)
    };
\addplot[
    color=cyan,
    mark=x,
    ]
    coordinates { 
(1,85.74)
(2,81.30)
(3,85.05)
(4,47.96)
(5,43.19)
    };
\nextgroupplot[xtick={1,...,5},xticklabels={16,32,64,128,256}]
\addplot[
    color=red,
    mark=x,
    ]
    coordinates {
  (1,72.32)
  (2,90.53)
  (3,96.97)
  (4,99.18)
  (5,99.61)
    };
\addplot[
    color=magenta,
    mark=x,
    ]
    coordinates {
  (1,47.21)
  (2,46.74)
  (3,46.51)
  (4,47.25)
  (5,46)
    };
\addplot[
    color=green,
    mark=x,
    ]
    coordinates {
  (1,42.19)
  (2,42.14)
  (3,42.16)
  (4,42.20)
  (5,42)
    };

\addplot[
color=blue,
mark=x,
]
coordinates {
(1,65.64)
(2,67.19)
(3,68.09)
(4,61.90)
(5,61.04)
};    
\addplot[
    color=orange,
    mark=x,
    ]
    coordinates { 
(1,70.20)
(2,67.94)
(3,64.29)
(4,61.55)
(5,69.52)
    };
\addplot[
    color=cyan,
    mark=x,
    ]
    coordinates { 
(1,75.67)
(2,95.99)
(3,84.38)
(4,62.05)
(5,61.36)
    };
\nextgroupplot[xtick={1,...,5},ylabel={Malfunction}, ytick pos=left,xticklabels={16,32,64,128,256}]
\addplot[
    color=red,
    mark=x,
    ]
    coordinates {
  (1,60.53)
  (2,82.22)
  (3,94.72) 
  (4,98.93)
  (5,99.22)
    };
\addplot[
    color=magenta,
    mark=x,
    ]
    coordinates {
  (1,19.398)
  (2,19.678)
  (3,19.69)
  (4,19.30)
  (5,19.40) 
    };
\addplot[
    color=green,
    mark=x,
    ]
    coordinates {
  (1,12.720)
  (2,12.583)
  (3,12.57)
  (4,12.62)
  (5,12.72) 
    };

\addplot[
    color=blue,
    mark=x,
    ]
    coordinates {
(1,60.93)
(2,51.47)
(3,47.51)
(4,46.73)
(5,50.85)
    };
\addplot[
color=orange,
mark=x,
]
coordinates {
(1,55.69)
(2,55.19)
(3,55.88)
(4,52.02)
(5,54.24)
};
\addplot[
    color=cyan,
    mark=x,
    ]
    coordinates { 
(1,87.38)
(2,48.80)
(3,46.95)
(4,46.73)
(5,47.12)
    };
\nextgroupplot[xlabel={$T$},xticklabels={16,32,64,128,256},xtick={1,...,5}]
\addplot[
    color=red,
    mark=x,
    ]
    coordinates {
  (1,70.49)
  (2,89.13)
  (3,98.31)
  (4,99.51)
  (5,99.17)
    };
\addplot[
    color=magenta,
    mark=x,
    ]
    coordinates {
 (1,26.008)
 (2,31.228)
 (3,31.28)
 (4,24.71)
 (5,26.01)
    };
\addplot[
    color=green,
    mark=x,
    ]
    coordinates {
  (1,13.266)
  (2,13.054)
  (3,13.06)
  (4,13.12)
  (5,13.27)
    };
\addplot[
    color=blue,
    mark=x,
    ]
    coordinates {
(1,61.64)
(2,62.04)
(3,54.73)
(4,50.66)
(5,55.06)
    };
\addplot[
    color=orange,
    mark=x,
    ]
    coordinates { 
(1,67.48)
(2,64.68)
(3,56.39)
(4,63.12)
(5,55.94)
    };
\addplot[
    color=cyan,
    mark=x,
    ]
    coordinates { 
(1,82.21)
(2,93.27)
(3,95.90)
(4,50.32)
(5,45.12)
    };
\nextgroupplot[xtick={1,...,5},legend style={at={($(0,0)-((3cm,1.5cm)$)},legend
        columns=6,fill=none,draw=black,anchor=center,align=center},xticklabels={16,32,64,128,256}]
\addplot[
    color=red,
    mark=x,
    ]
    coordinates { 
  (1,71.49)
  (2,91.13)
  (3,98.34)
  (4,99.52)
  (5,99.27)
    };
\addplot[
    color=magenta,
    mark=x,
    ]
    coordinates {
  (1,29.418)
  (2,36.197)
  (3,31.23)
  (4,30.53)
  (5,30)
    };
   \addplot[
   color=green,
    mark=x,]  coordinates {
  (1,21.014)
  (2,21.016)
  (3,21.03)
  (4,21.10)
  (5,21)
};
\addplot[
    color=blue,
    mark=x,
    ]
    coordinates {
(1,52.05)
(2,51.76)
(3,52.08)
(4,51.87)
(5,52.15)
    };
\addplot[
    color=orange,
    mark=x,
    ]
    coordinates { 
(1,50.95)
(2,52.57)
(3,51.74)
(4,51.86)
(5,58.72)
    };
\addplot[
    color=cyan,
    mark=x,
    ]
    coordinates { 
(1,52.03)
(2,51.94)
(3,51.77)
(4,51.92)
(5,52.16)
    };
\addlegendentry{CAN-BERT};  
\addlegendentry{iForest}; 
\addlegendentry{PCA}; 
\addlegendentry{BiLSTM-AE-4}; 
\addlegendentry{LSTM-AE-4}; 
\addlegendentry{LSTM-AE-8}; 
\end{groupplot}
\end{tikzpicture}
\caption{Comparision of the CAN-BERT model with other anomaly detection baselines using the F1-score percentage metric, for different message injection attacks applied on different car models w.r.t sequence length $T$.}
\label{fig:models_comparision}
\end{figure*}
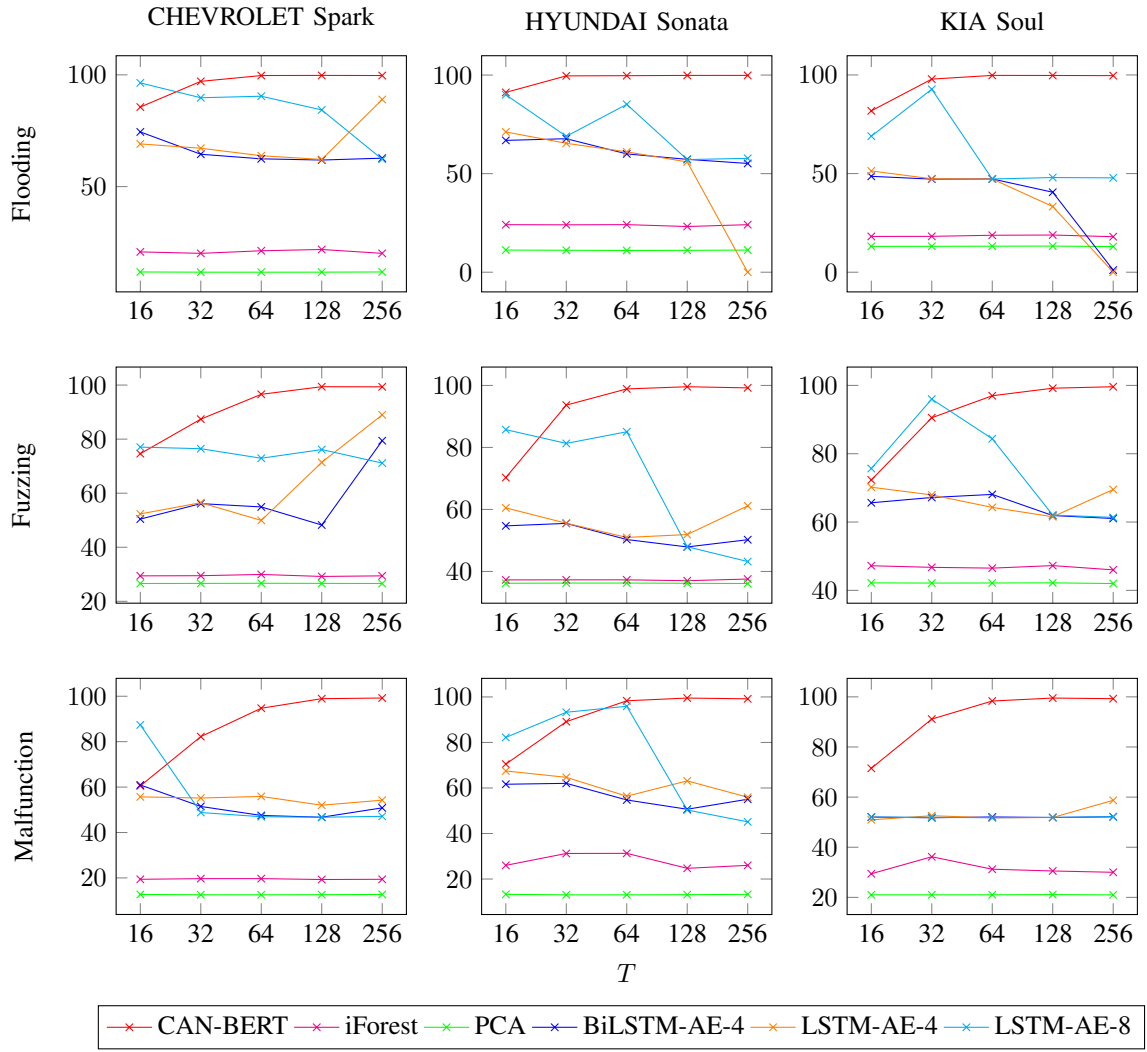

\section{Results}
\label{s5}
To evaluate our model, we leverage the Python deep learning framework Pytorch \cite{pytorch}. We train and evaluate them on NVIDIA® Tesla® V100S with 32 GB HBM2 memory. 

\subsection{Model Configuration \& Hyperparameter tuning}

As presented in Table \ref{tab:model_hyperparams}, for CAN-BERT, we have chosen the total number of transformer encoder layers as 4. In each transformer layer, the position-wise feed forward network is composed of two dense layers where the first one projects $d$=256 dimensional CAN identifier embedding into $d_{ff}$=512 dimensional space, followed by a ReLU activation. The second dense layer maps back the 512-dimensional vector into the $d$-dimensional space. 
\begin{table}[h]
    \centering
    \caption{CAN-BERT model configuration}
    \begin{tabular}{|c|c|}
    \hline
         Parameter& Value \\
         \hline
          \textit{N} & 4\\
         \hline
         $d_{model}$ & 256\\
         \hline
         $d_{ff}$ & 512\\
         \hline
         \textit{h} & 1\\
         \hline
         $P_{drop}$ & 0.1\\
         \hline
         $m$ & 0.45\\
         \hline
         $\#$ Candidates & 5 \\
         \hline
         Optimizer & Adam\\
         \hline
         Adam $\beta_{1}$ & 0.9\\
         \hline
         Adam $\beta_{2}$ & 0.999 \\
         \hline
          Learning rate & 0.001\\
         \hline
         Batch size & 32\\
         \hline
         $\#$ Epochs & 200\\
         \hline
         Patience & 10\\
         \hline
        
    \end{tabular}
    \label{tab:model_hyperparams}
\end{table}
For training, we use a batch size of 32, a learning rate of 0.001 and the Adam optimizer \cite{Adam} with its default parameters $\beta_{1}=0.9$ and $\beta_{2}=0.999$. To avoid overfitting, we employ the same dropout of $P_ {drop}$=0.1 for all dropout layers in our network. Moreover, we apply early stopping for a total number of 200 epochs and a patience of 10 epochs as a form of regularization used to avoid overfitting. 

The hyperparameters, including the ratio of masked CAN IDs in a sequence $m$, and $h$ the number of attention heads are tuned based on a validation set for the three car types and the different message injection attacks. As seen in Figure \ref{hyperparameters}, raising the ratios of masked CAN IDs in the sequences from 0.1 to 0.45 somewhat improves F1 scores, however increasing the ratios further degrades performance, as is the case for $m$=0.65. Furthermore, the model performance is relatively stable by setting different attention head $h \in \{1,2,4,8\}$ values for each mask ratio $m \in \{0.15,0.3,0.45,0.6\}$. Therefore, in our in-vehicle intrusion detection use case, a single attention head is sufficiant to detect different types of intrusions.
Note that, in our experiments, we use the same ratio of masked CAN IDS $m$=0.45 and $h$=1 for both training and inference phases.

\subsection{Model Accuracy}
As seen in Figure \ref{fig:models_comparision}, we compare performance of the CAN-BERT model with other baselines approaches for different sequence length $T$ using the F1-score metric. In fact, we varied the sequence length among values of 16, 32, 64, 128 and 256 CAN IDs per sequence in the experiments. If our approaches could identify a message injection attack in a shorter sequence length, it would be more advantageous in a practical situation. The traditional machine learning algorithms such as Isolation Forest (iForest), and Principal Component Analysis (PCA) perform poorly and maintain the same F1-score metric w.r.t sequence length. Because these models presume small datasets with a limited number of features, they fail to discover abnormalities in  high-dimensional datasets. Because of this, a significant fraction of irrelevant features may effectively disguise the underlying abnormalities in the input data, resulting in poor anomaly identification performance when dealing with large input dimensions. Meanwhile, both deep learning based models autoencoder (AE) and CAN-BERT outperformed the traditional anomaly detection models over different window sizes. However, when the length of the CAN ID sequence is increased, both models performed  oppositely. In contrast to the baseline models, our suggested model, CAN-BERT, significantly outperforms them by huge margins and obtains respectable F1 scores $\in [0.85,0.99]$, demonstrating the usefulness of using BERT-based models to capture the patterns of CAN ID sequences when $T  \geq 32$. However, on short sequence length as is the case for $T=16$, the model performs modestly with F1-score $\in [0.6,0.9]$ for different kind of attacks. These experiments, therefore, reveal that by using self-supervised training tasks, CAN-BERT can effectively model medium to long normal CAN ID sequences and accurately detect anomalous sequences.

\subsection{Model Complexity}
From a practical point of view, we must assess not only the classification performance but also the model complexity to check if the model's ability for real-time in-vehicle intrusion detection in CAN networks. Therefore, we assessed the inference time per sample as well as the number of parameters for the CAN-BERT model w.r.t different car types.
\begin{table}[h]
    \centering
    \caption{CAN-BERT model complexity}
    \begin{tabular}{|c|c|c|}
    \hline
        Vehicle& Features & Values\\
         \hline
        \multirow{3}{*}{CHEVROLET Spark}&Number of Parameters & 2,937,422 \\
        \cline{2-3}
        &Model Size (MB) & $[20,70]$\\
         \cline{2-3}
         &Inference Time (ms) & $[0.8,3.1]$\\
         \hline
        \multirow{3}{*}{HYUNDAI Sonata}&Number of Parameters & 3,149,291\\
     \cline{2-3}
             &Model Size (MB) & $[20,70]$\\
         \cline{2-3}
     &Inference Time (ms)& $[0.8,3.5]$\\
     \hline
        \multirow{3}{*}{KIA Soul} &Number of Parameters & 3,163,142\\
         \cline{2-3}
                 &Model Size (MB) & $[20,70]$\\
         \cline{2-3}
         &Inference Time (ms)& $[0.8,3.8]$\\
         \hline
    \end{tabular}
    \label{tab:model_complexity}
\end{table}
As depicted in Table \ref{tab:model_complexity}, the intrusion detection inference time varies between 0.8 and 3.1 ms w.r.t CAN ID sequence length. Hence, when considering a sequence length of 32 CAN IDs, our model detects an intrusion in 0.9 to 1 ms, which is suitable for real-time detection. Furthermore, having a size between 20MB and 70 MB and a number of parameters ranging between 2 to 3 millions, our model can be either deployed in performant ECU or even on a cloud server wirelessly connected to the vehicle.

\section{Conclusion}
\label{s6}
Identification of intrusions within the vehicle is critical for defending it against malicious cyberattacks. In this paper, we suggest CAN-BERT, a self-supervised intrusion detection system based on BERT model, for in-vehicle intrusion detection. Experimental results on benchmark datasets for different CAN ID sequence length have shown that CAN-BERT surpasses state-of-the-art techniques for CAN ID sequence anomaly detection with an F1-score ranging between 0.81 and 0.99 for different type of attacks and is appropriate for real-time detection with an inference time ranging between 0.8 and 3 ms w.r.t CAN ID sequence length. For future work, we aim to deploy our model on embedded electronic control units (ECU) and test the model efficiency in a real vehicle environment.


\begin{thebibliography}{}
\bibitem{AutoEth_book}Matheus, Kirsten, and Thomas Königseder. Automotive ethernet. Cambridge University Press, 2021.
\bibitem{BERT}Devlin, Jacob, et al. "Bert: Pre-training of deep bidirectional transformers for language understanding." arXiv preprint arXiv:1810.04805 (2018).
\bibitem{Transformer}Vaswani, Ashish, et al. "Attention is all you need." Advances in neural information processing systems 30 (2017).
\bibitem{Normalization}Ba, Jimmy Lei, Jamie Ryan Kiros, and Geoffrey E. Hinton. "Layer normalization." arXiv preprint arXiv:1607.06450 (2016).
\bibitem{Bi-GPT}Nam, Minki, Seungyoung Park, and Duk Soo Kim. "Intrusion detection method using bi-directional GPT for in-vehicle controller area networks." IEEE Access 9 (2021): 124931-124944.
\bibitem{Language_anomaly_bert}Li, Bai, et al. "How is BERT surprised? Layerwise detection of linguistic anomalies." arXiv preprint arXiv:2105.07452 (2021).
\bibitem{LogBERT}Guo, Haixuan, Shuhan Yuan, and Xintao Wu. "Logbert: Log anomaly detection via bert." 2021 International Joint Conference on Neural Networks (IJCNN). IEEE, 2021.
\bibitem{LAnoBERT}Lee, Yukyung, Jina Kim, and Pilsung Kang. "LAnoBERT: System Log Anomaly Detection based on BERT Masked Language Model." arXiv preprint arXiv:2111.09564 (2021).
\bibitem{NeuralLog}Le, Van-Hoang, and Hongyu Zhang. "Log-based anomaly detection without log parsing." 2021 36th IEEE/ACM International Conference on Automated Software Engineering (ASE). IEEE, 2021.
\bibitem{IOT_BERT}Yu, Keping, et al. "Securing critical infrastructures: Deep-Learning-Based threat detection in IIoT." IEEE Communications Magazine 59.10 (2021): 76-82.
\bibitem{TS-BERT}Dang, Weixia, et al. "TS-Bert: Time Series Anomaly Detection via Pre-training Model Bert." International Conference on Computational Science. Springer, Cham, 2021.
\bibitem{AE}Vincent, Pascal, et al. "Extracting and composing robust features with denoising autoencoders." Proceedings of the 25th international conference on Machine learning. 2008.
\bibitem{Dataset}Hyunjae Kang, Byung Il Kwak, Young Hun Lee, Haneol Lee, Hwejae Lee, Huy Kang Kim, February 3, 2021, "Car Hacking: Attack \& Defense Challenge 2020 Dataset", IEEE Dataport, doi: \url{https://dx.doi.org/10.21227/qvr7-n418}
\bibitem{PCA}Abdi, Hervé, and Lynne J. Williams. "Principal component analysis." Wiley interdisciplinary reviews: computational statistics 2.4 (2010): 433-459.
\bibitem{IsolationForest}Liu, Fei Tony, Ting, Kai Ming and Zhou, Zhi-Hua. “Isolation-based anomaly detection.” ACM Transactions on Knowledge Discovery from Data (TKDD) 6.1 (2012): 3.
\bibitem{AE1}Rumelhart, David E., Geoffrey E. Hinton, and Ronald J. Williams. "Learning representations by back-propagating errors." nature 323.6088 (1986): 533-536.
\bibitem{pytorch}Pytorch framework. \url{https://pytorch.org/.}
\bibitem{Adam}Kingma, Diederik P., and Jimmy Ba. "Adam: A method for stochastic optimization." arXiv preprint arXiv:1412.6980 (2014).
\bibitem{canet}Hanselmann, Markus, et al. "CANet: An unsupervised intrusion detection system for high dimensional CAN bus data." Ieee Access 8 (2020): 58194-58205.
\bibitem{cnn}Song, Hyun Min, Jiyoung Woo, and Huy Kang Kim. "In-vehicle network intrusion detection using deep convolutional neural network." Vehicular Communications 21 (2020): 100198.
\end{thebibliography}
\end{document}